\documentclass[letterpaper]{article} 
\usepackage{aaai2026}  
\usepackage{times}  
\usepackage{helvet}  
\usepackage{courier}  
\usepackage[hyphens]{url}  
\usepackage{graphicx} 
\usepackage{natbib}  
\usepackage{caption} 
\usepackage{marvosym} 
\usepackage{algorithm}
\usepackage{algorithmic}
\usepackage{newfloat}
\usepackage{listings}
\usepackage{makecell}
\usepackage[T1]{fontenc}
\usepackage{multicol}
\usepackage{multirow}
\usepackage{amssymb}
\usepackage{xcolor} 
\usepackage{graphicx}
\usepackage{arydshln} 
\usepackage{enumitem}
\usepackage{amsmath}
\usepackage{booktabs}

\usepackage{graphicx}
\usepackage{subcaption}
\usepackage{diagbox}
\usepackage{newfloat}
\usepackage{listings}
\DeclareCaptionStyle{ruled}{labelfont=normalfont,labelsep=colon,strut=off} 
\floatstyle{ruled}
\newfloat{listing}{tb}{lst}{}
\floatname{listing}{Listing}
\title{Unleashing the Power of Image-Tabular Self-Supervised Learning \\ via Breaking Cross-Tabular Barriers}
\author{
    Yibing Fu\textsuperscript{\rm 1},
    Yunpeng Zhao\textsuperscript{\rm 1},
    Zhitao Zeng\textsuperscript{\rm 1},
    Cheng Chen\textsuperscript{\rm 3,4},
    Yueming Jin\textsuperscript{\rm 1,2,}\thanks{Yueming Jin is the corresponding author}
    }
\affiliations{
    \textsuperscript{\rm 1}Department of Biomedical Engineering, National University of Singapore\\
    \textsuperscript{\rm 2}Department of Electrical and Computer Engineering, National University of Singapore\\
    \textsuperscript{\rm 3}Department of Electrical and Electronic Engineering, The University of Hong Kong \\
    \textsuperscript{\rm 4}School of Biomedical Engineering, The University of Hong Kong\\
    \{yibingfu, yunpeng.zhao\}@u.nus.edu, cchen@eee.hku.hk, \{zhitao, ymjin\}@nus.edu.sg

}

\usepackage{bibentry}

\begin{document}

\maketitle

\begin{abstract}
Multi-modal learning integrating medical images and tabular data has significantly advanced clinical decision-making in recent years. Self-Supervised Learning (SSL) has emerged as a powerful paradigm for pretraining these models on large-scale unlabeled image-tabular data, aiming to learn discriminative representations. However, existing SSL methods for image-tabular representation learning are often confined to specific data cohorts, mainly due to their rigid tabular modeling mechanisms when modeling heterogeneous tabular data. This inter-tabular barrier hinders the multi-modal SSL methods from effectively learning transferrable medical knowledge shared across diverse cohorts.
In this paper, we propose a novel SSL framework, namely CITab, designed to learn powerful multi-modal feature representations in a cross-tabular manner. We design the tabular modeling mechanism from a semantic-awareness perspective by integrating column headers as semantic cues, which facilitates transferrable knowledge learning and the scalability in utilizing multiple data sources for pretraining. Additionally, we propose a prototype-guided mixture-of-linear layer (P-MoLin) module for tabular feature specialization, empowering the model to effectively handle the heterogeneity of
tabular data and explore the underlying medical concepts. We conduct comprehensive evaluations on Alzheimer's disease diagnosis task across three publicly available data cohorts containing 4,461 subjects. Experimental results demonstrate that CITab outperforms state-of-the-art approaches, paving the way for effective and scalable cross-tabular multi-modal learning.
\end{abstract}

\begin{links}
    \link{Code}{https://github.com/jinlab-imvr/CITab}
\end{links}

\section{Introduction}

\begin{figure}[!ht]
\centering
\includegraphics[width=0.9\linewidth]{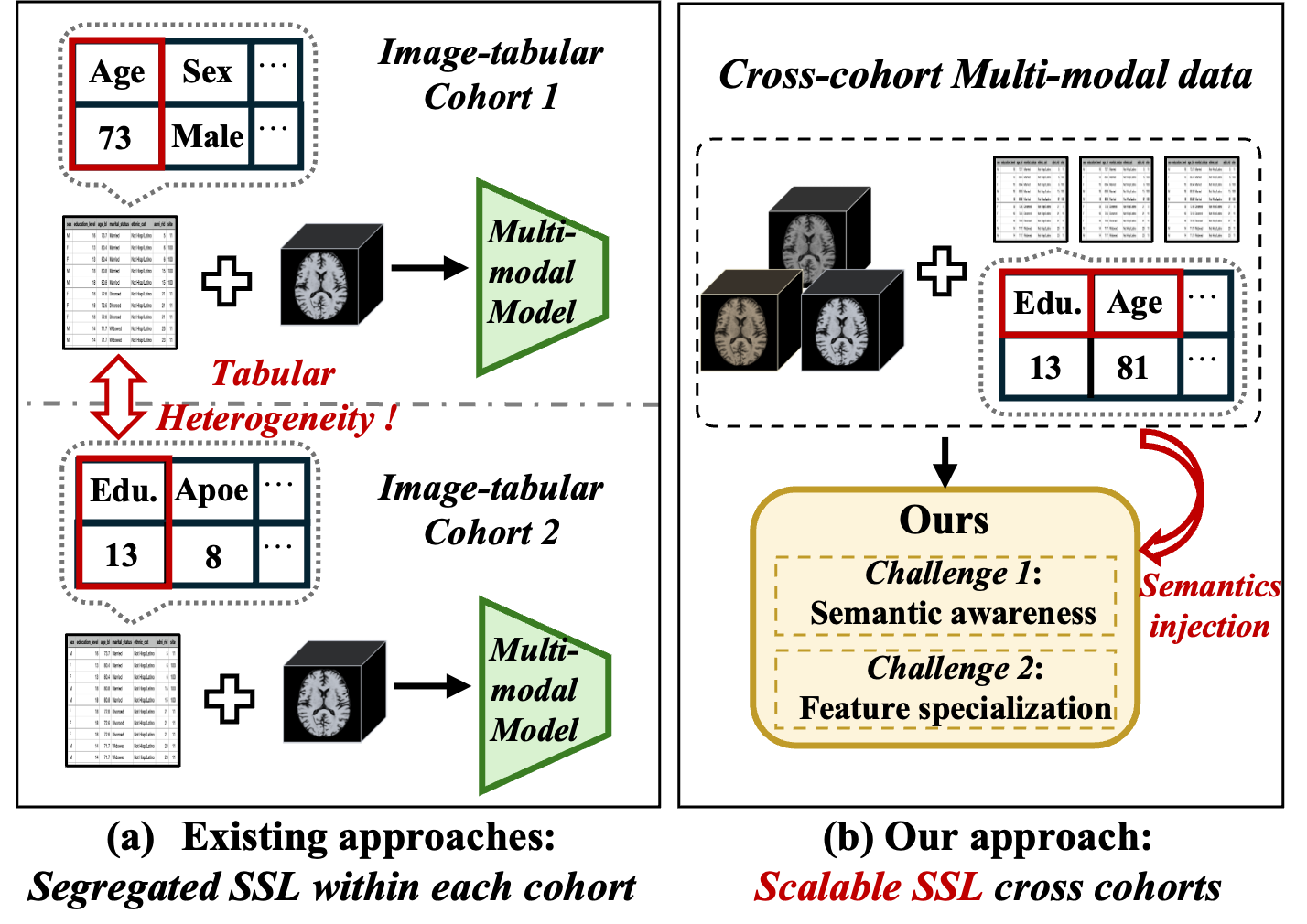}
\caption{Comparison between existing and our proposed methods for image-tabular SSL across multiple cohorts. (a) Existing approaches perform SSL and fine-tuning over each cohort separately due to tabular heterogeneity, which leads to under-exploration of shared medical knowledge.(b) Our proposed CITab framework enables scalable SSL in cross-tabular scenarios, fostering the generation of robust multi-modal representations.}
\label{fig1}
\end{figure}

Medical images and tabular data are two fundamental modalities in modern healthcare systems \cite{bycroft2018ukbb,du2024tip,Hager_2023_CVPR}. Tabular data is typically collected alongside imaging data, providing complementary clinical information such as demographic details and laboratory tests \cite{du2024tip}. Integrating these two modalities enables a more comprehensive assessment of a patient’s condition, which is crucial for real-world clinical decision-making \cite{bai2020population,buntin2011benefits,chaudhry2006systematic, wang2023multi, fu2023deep}. With the advancement of deep learning technique, several studies have demonstrated the benefits of integrating imaging and tabular data to enhance diagnostic performance \cite{vale2021multisurv,liu2023functional,wolf2022daft,dong2025multimodal}. 

However, these attempts often adopt a supervised-learning paradigm and require expensive human annotation for training, which impedes their scaling. Recently, self-supervised learning (SSL) has emerged as a powerful paradigm to alleviate this issue by learning from massive unannotated data via some pretext tasks\cite{gui2024survey}, which significantly enhances the  power of multi-modal models. Although these SSL-based methods have made substantial progress within the healthcare field \cite{wang2022multim, zhao2024dealing}, research incorporating tabular data and medical images for cross-modality SSL remains relatively scarce. Recently, two studies have started exploring this direction \cite{Hager_2023_CVPR,du2024tip}. Hager \textit{et al.} \cite{Hager_2023_CVPR} proposed MMCL, which jointly trains image and tabular encoders through  contrastive learning. Similarly, Du \textit{et al.} \cite{du2024tip} designed a SSL strategy to handle missing tabular values.

Although these two methods provide early attempts into image-tabular representation learning, they are limited to performing SSL on a specific image-tabular cohort and lack scalability to leveraging multiple image-tabular data cohorts for pretraining. As shown in Fig.\ref{fig1}-(a), tables from distinct datasets often differ drastically in their structures. For instance, the number and order of columns always vary significantly across tables. Current methods \cite{Hager_2023_CVPR,du2024tip} cannot overcome such tabular heterogeneity, which hinders them to perform pretraining across diverse tables. This limitation primarily stems from their rigid tabular modeling mechanisms, which do not accommodate tables with varying structures. For example, TIP \cite{du2024tip} aligns its tabular embedding and position encoding with the number and order of columns in the input table. This limitation significantly restricts the pretraining data scale, thereby compromising the capture of transferable knowledge across different multi-modal cohorts.

Consequently, it becomes a critical yet challenging task to design \textbf{a scalable and flexible image-tabular SSL framework}, enabling effective cross-tabular learning. As aforementioned, the disorganized column positions pose a challenge for existing models, which relies on fixed column structures by projecting cell values rather than capturing underlying semantics which limits their scalability to heterogeneous tables. Therefore, it is essential to empower the model to \textbf{capture underlying semantics} regardless of the disordered column arrangements when encountering heterogeneous tables. Moreover, during cross-tabular learning, the model is exposed to a wide variety of tabular attributes. Unique attributes from distinct tables require careful consideration to ensure effective tabular feature specialization. Robust cross-table learning hinges on adapting to heterogeneous features caused by variations in structure and semantics. However, naive applications of transformer-based \cite{du2024tip} or MLP-based \cite{Hager_2023_CVPR} often struggle to accommodate such variations. This underscores the need for a \textbf{dynamic feature calibration mechanism} that adeptly captures table heterogeneity during cross-tabular learning.

To address these challenges, we propose a novel framework for \underline{c}ross-tabular \underline{i}mage-\underline{tab}ular pretraining (\textbf{CITab}), which set the first attempt to explore medical images and tabular SSL in cross-tabular scenarios. Subsequently, considering column semantics can provide crucial insights, helping the model recognize underlying column patterns and adapt to varying column arrangements, we design a semantic-aware tabular-embedding mechanism. Specifically,
we leverage linguistic information, i.e., header names, as semantic clues for distinct columns and design a semantic-aware embedding mechanism. By integrating these semantic cues with table values, the method can eliminate the reliance on position encoding constrained by a fixed number and order of columns, enabling flexible and scalable cross-table learning. Furthermore, inspired by the mixture-of-experts framework \cite{fedus2022switch}, we propose a prototype-guided mixture-of-linear (P-MoLin) module, which employs learnable prototypes as feature anchors to capture diverse underlying patterns and dynamically calibrate tabular features across heterogeneous tables, thereby enhancing both flexibility and effectiveness during cross-tabular learning. In summary, our main contribution can be summarized as follows:

\begin{itemize}
    \item To the best of our knowledge, we set the first attempt to explore image-tabular SSL in cross-tabular scenarios, paving the way for scaling image-tabular SSL methods.
    \item We design a semantic-aware tabular-embedding mechanism, which empowers the model to recognize underlying column patterns and adapt to varying column orders.
    \item We propose a P-MoLin module, facilitating dynamic feature calibration when modeling heterogeneous tables, enhancing the  flexibility during cross-tabular learning. 
    \item We comprehensively evaluate our method over three data cohorts encompassing 4,461 3D brain magnetic resonance images (MRIs) and corresponding tabular data over the task of Alzheimer’s disease diagnosis, demonstrating superior performance of our proposed method.
\end{itemize}

\section{Related Work}

With the advancement of deep learning technique, several studies have demonstrated the benefits of integrating imaging and tabular data to enhance diagnostic performance \cite{vale2021multisurv,liu2023functional,wolf2022daft,dong2025multimodal}. For instance, Luis \textit{et al.} incorporated imaging and tabular information to improve cancer survival predictions \cite{vale2021multisurv}. 
However, simply appending tabular features to image representations often yields limited improvements due to insufficient cross-modal interaction. To address this, Wolf \textit{et al.} proposed DAFT, which conditions convolutional feature maps on both image and tabular inputs, facilitating bidirectional information exchange through an auxiliary neural network \cite{wolf2022daft}. More recently, Hyperfusion ~\cite{duenias2024hyperfusion} introduced hypernetworks to dynamically generate parameters for fusing imaging and clinical data, demonstrating effectiveness in  Alzheimer's disease prediction.

However, the above multi-modal fusion approaches are designed in a supervised manner, which require extensive annotated data. To mitigate the annotation cost, self-supervised learning (SSL) emerged for learning multi-modal representations from unlabeled data. MMCL pioneered image-tabular SSL in healthcare using contrastive learning to jointly train encoders~\cite{Hager_2023_CVPR}. Similarly, Huang \textit{et al.} leveraged contrastive learning combined with attention mechanisms on tabular features to enhance Alzheimer's disease prediction \cite{huang2023multimodal}. Further extending the contrastive SSL paradigms, 
Du \textit{et al.} proposed TIP, which introduced a transformer-based tabular encoder inherently resilient to missing data and employed SSL objectives like masked tabular reconstruction and image-tabular matching during pretraining~\cite{du2024tip}. Despite these advancements, these multi-modal SSL methods are confined to single datasets due to rigid tabular modeling that cannot handle structural heterogeneity, hindering scalable pretraining on diverse multi-modal cohorts.

\section{Methodology}

\begin{figure*}[!ht]
\centering
\includegraphics[width=0.77\linewidth]{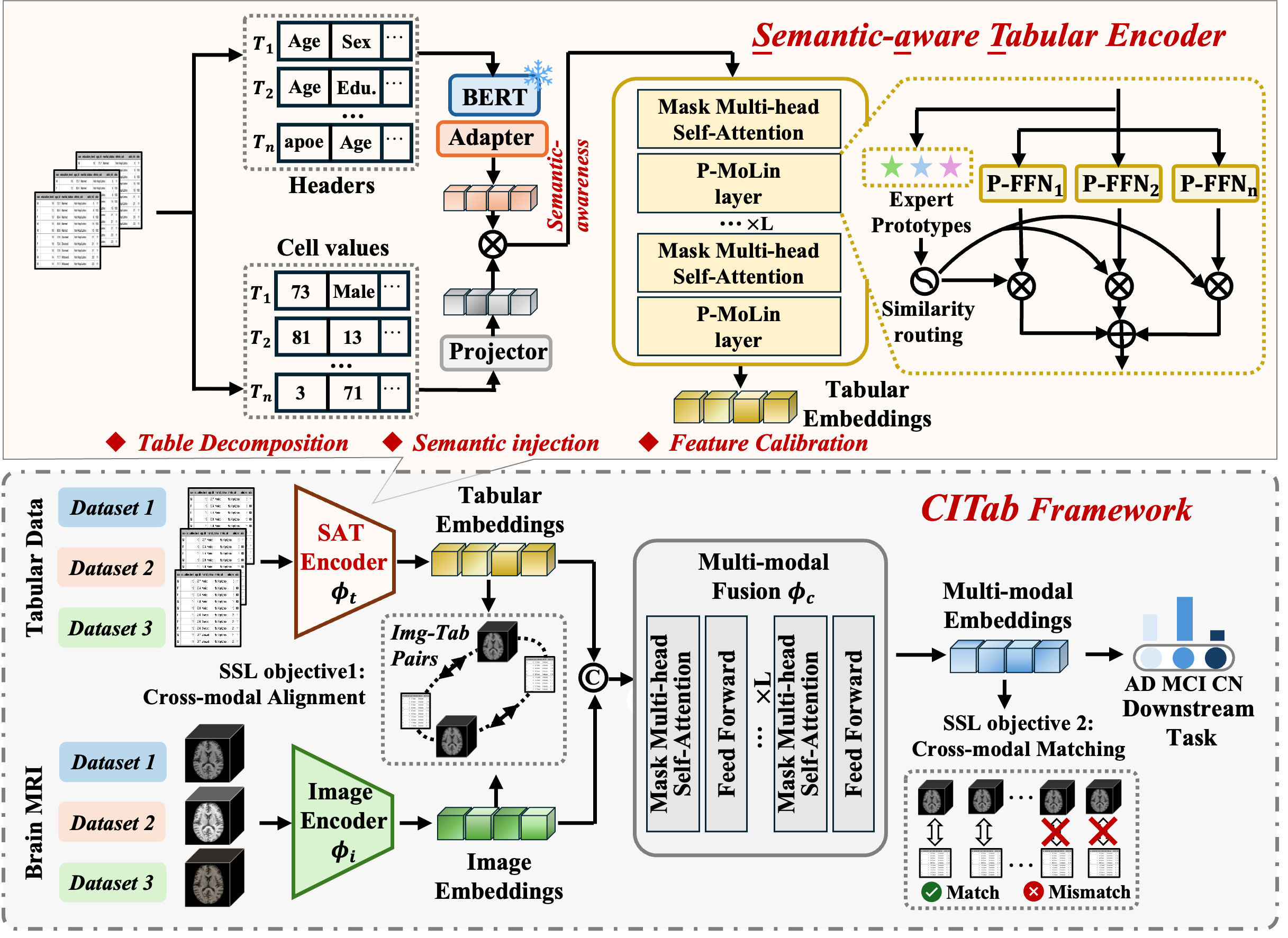}
\caption{The framework of the proposed CITab. The tailored SAT Encoder effectively handles tabular heterogeneity via semantic-awareness embedding and prototype-guided feature calibration. The obtained tabular embeddings are fused with image embeddings through a multi-modal fusion module, enabling scalable cross-tabular pretraining.}
\label{framework}
\end{figure*}

In this section, we introduce CITab, a novel image-tabular SSL framework which is tailored for cross-tabular learning across distinct data cohorts. To effectively model heterogeneous tabular data, we propose a \underline{s}emantic-\underline{a}ware \underline{t}abular (SAT) encoder, which enhances semantic-awareness via incorporating linguistic information, i.e., column header, to boost transferable medical knowledge learning. Additionally, we introduce a P-MoLin module for effective feature specialization. Finally, we present an SSL strategy to improve image-tabular representation learning. The overall framework is illustrated in Fig.\ref{framework}.

\subsection{Overall Framework of CITab}

As depicted in Fig.\ref{framework}, our CITab takes a 3D-MRI and corresponding tabular data as input. During the SSL stage, we conduct scalable pretraining over all available data cohorts in a cross-tabular manner. Given an image $\mathbf{I}$ and its paired tabular data $\mathbf{T}$ from an available cohort, we introduce an image encoder $\mathbf{\Phi}_i$, which employs 3D-ResNet50 \cite{he2016deep} with a projection layer for dimension transformation as the backbone, yielding the image feature embedding $\mathbf{f}_i=\mathbf{\Phi}_i(\mathbf{I})\in\mathbf{R}^{C_i \times D}$. $D$ and $C_i$ denote the embedding dimension and the length of image embedding sequence respectively. Subsequently, we propose a tailored SAT encoder $\mathbf{{\Phi}}_t$ to obtain the tabular feature embedding  $\mathbf{f}_t=\mathbf{\Phi}_t(\mathbf{T})\in\mathbf{R}^{(C_t + 1) \times D}$. Note that $C_t$ denote the length of tabular embedding sequence which equals to the number of table columns, and the additional token signifies an appended class token designed to capture global information. 

After obtaining the embeddings $\mathbf{f}_i$ and $\mathbf{f}_t$, we concatenate them along the sequence dimension to construct a unified representation $\mathbf{f}_{con}\in\mathbf{R}^{(C_i + C_t + 1) \times D}$. To integrate multi-modal information, we propose a unified transformer-based fusion module $\mathbf{\Phi}_c$, consisting of stacked self-attention layers that produce the multi-modal embedding $\mathbf{f}_c$. Unlike prior works such as \cite{du2024tip}, which adopt cross-attention mechanisms to model inter-modal relationships, we adopt a unified self-attention design based on its inherent ability to accommodate input sequences of varying lengths, making it particularly well-suited for modeling heterogeneous tabular data with differing numbers of columns. 

Additionally, the varying number of columns poses challenges for efficient parallel training. To mitigate this, we adopt a padding strategy inspired by practices in large-scale language models such as BERT \cite{devlin2019bert}. Specifically, each tabular embedding is padded to a predefined maximum length with a mask token. During training, we apply masked self-attention within both the tabular encoder $\mathbf{\Phi}_t$ and the fusion module $\mathbf{\Phi}_c$, ensuring that the attention mechanism operates only over valid (non-padded) tabular tokens.

Finally, these three yielding embeddings, i.e., $\mathbf{f}_i$, $\mathbf{f}_t$ and $\mathbf{f}_c$, are leveraged in two pretext SSL objectives, i.e., cross-model alignment and cross-modal matching, during the cross-tabular pretraining stage. During following fine-tuning stage, we feed the class token in $\mathbf{f}_c$ into a linear layer, which aims to perform the multi-modal classification task supervised by cross-entropy loss.

\subsection{Semantic-aware Tabular Embedding}

To effectively handle the semantic ambiguity arising from disordered column positions in cross-tabular scenarios, we design a semantic-aware tabular embedding mechanism that leverages column headers as semantic clues, as shown in Fig.\ref{framework}. The underlying mechanism can be conceptually divided into table decomposition and semantic-awareness injection. Basically, a tabular sample, i.e., an arbitrary row in the table, can be represented as a collection of structured data $\mathbf{v} \in \mathbf{R}^{C_t}$, each associated with a corresponding column header $\{h_i\}_{i=1}^{C_t}$, where $C_t$ denotes the number of columns.

Given such a tabular sample, we apply a shared linear projection layer across columns to map each cell value $v \in \mathbf{R}$ into a high-dimension embedding space. As a result, we obtain a value embedding $\mathbf{v} \in \mathbf{R}^{C_t \times D}$, where $D$ denotes the embedding dimension. For the corresponding column headers, we leverage a pretrained language model, BERT \cite{devlin2018bert}, to extract contextualized sentence embeddings $\mathbf{h}_i \in \mathbf{R}^{D_h}$ for each header $h_i$, with the model parameters kept frozen during training. All header embeddings are then stacked to form the header embedding $\mathbf{h} \in \mathbf{R}^{C_t \times D_h}$, which captures the underlying semantics of the columns.

Subsequently, we feed $\mathbf{h}$ to an linear adapter layer to perform dimension alignment, yielding the aligned header embedding $\hat{\mathbf{h}}$. Finally, these two obtained features are fused to generate the semantic-aware tabular embedding $\mathbf{\hat{f}}_{t} = \mathbf{v}\odot\hat{\mathbf{h}}$. $\odot$ refers to element-wise multiplication. As a result, in contrast to methods such as \cite{du2024tip}, which rely on fixed positional embeddings to model tabular data, our approach empowers the model to capture underlying semantics without being constrained by column arrangements, thereby overcoming inter-tabular structural barriers and enhancing adaptability across heterogeneous tabular data.

\subsection{Tabular Feature Calibration}

During cross-tabular learning, the model is exposed to a wide variety of tabular attributes, and the unique structure of each table necessitates dynamic adaptation to effectively capture its underlying semantics. Despite this heterogeneity, some attributes may implicitly reflect shared latent concepts. For instance, results from different cognitive tests can all indicate an individual's cognitive capacity. To leverage this property, as illustrated in Fig.\ref{framework}, we propose a P-MoLin module  that dynamically adapts to diverse tabular features while capturing shared latent concepts.

Within the SAT encoder, after obtaining the semantic-aware tabular embedding $\mathbf{\hat{f}}_{t}$, we employ a tailored tabular encoder $\mathbf{\Phi}_t$ for feature interactions. The proposed $\mathbf{\Phi}_t$ is composed of stacked masked multi-head self-attention (MSA) layers. Unlike standard transformers where each MSA layer is followed by a vanilla feed-forward network (FFN), we replace the FFN with our proposed P-MoLin module. Specifically, the P-MoLin module introduces $E$ learnable expert prototypes $\mathbf{P}\in\mathbf{R}^{E \times D}$, where each prototype serves to capture a distinct underlying concept. These prototypes serve as feature anchors, guiding tabular features toward specialized expert pathways according to their conceptual similarity.

Let $\mathbf{f}_{in}\in\mathbf{R}^{(C_t + 1) \times D}$ and $\mathbf{f}_{out}\in\mathbf{R}^{(C_t + 1) \times D}$ denote the input and output features of a given P-MoLin module. The input $\mathbf{f}_{in}$ is passed through $E$ prototype-guided feed-forward networks (P-FFNs) to generate expert-specific features. Simultaneously, we compute the similarity matrix $M = \mathbf{f}_{in} \cdot \mathbf{P}^T \in\mathbf{R}^{(C_t + 1) \times E}$, which measures the alignment between the input features and each prototype. To enable dynamic adaptation, $M$ is normalized using a softmax function along the expert dimension, producing routing weights that assign each token to the most relevant expert (P-FFN module), based on its conceptual alignment with the prototypes. This procedure is mathematically formulated as:
\begin{align}
\mathbf{f}_{out} = \sum_{e=1}^E \left( \frac{\exp(M_{:,e})}{\sum_{k=1}^{E} \exp(M_{:,k})} \right) \cdot \mathrm{FFN}_e (\mathbf{f}_{in}),
\end{align}
where $M_{:,e}\in\mathbf{R}^{C_t+1}$ is the $e$-\textit{th} column of $M$, and $\mathrm{FFN}_e$ is the $e$-\textit{th} P-FFN module. We empirically set $E$ to 5 in our default setting. In summary, by dynamically adapting tabular features, P-MoLin enhances the model's expressiveness, effectively addressing the heterogeneity in tabular data.

\subsection{Self-supervised Learning Objective}

To enable the model to learn powerful multi-modal representations, we pretrain it using two pretext tasks. Firstly, we extract the class token in  $\mathbf{f}_t$ and apply spatial pooling to the image features $\mathbf{f}_i$ to obtain global representations for both modalities. We introduce two linear layers to get projected image and tabular embeddings $\hat{\mathbf{f}}_i$ and $\hat{\mathbf{f}}_t$ for similarity calculation. Subsequently, we exploit an image-tabular contrastive learning objective $\mathcal{L}_{itc}$ to align the image and tabular feature spaces, which can be formulated as:
\begin{equation}
\mathcal{L}_{itc} = 
 -\frac{1}{2B} \sum_{b=1}^{B} \left( \text{sim}({\hat{\mathbf{f}}}_{i,b}, {\hat{\mathbf{f}}}_{t,b}) + \text{sim}({\hat{\mathbf{f}}}_{t,b}, {\hat{\mathbf{f}}}_{i,b}) \right),
\end{equation}
\begin{equation}
\text{sim}({\hat{\mathbf{f}}}_{i,b}, {\hat{\mathbf{f}}}_{t,b}) = 
 \log \frac{\exp \left( {\hat{\mathbf{f}}}_{i,b}^\top {\hat{\mathbf{f}}}_{t,b} / \tau \right)}{\sum\limits_{k=1}^{B} \exp \left( {\hat{\mathbf{f}}}_{i,b}^\top \hat{\mathbf{f}}_{t,k} / \tau \right)},
\end{equation}
where $B$ and $\tau$ denote the batch size and temperature hyperparameter. 

To enhance cross-modal learning, we introduce an image-tabular matching task to predict whether given imaging and tabular inputs originate from the same sample. As depicted in Fig.\ref{framework}, positive pairs are formed by aligning image and tabular embeddings from the same sample. For negative pairs, following \cite{du2024tip}, we replace either the image or tabular embedding with the most similar embedding from the same batch, based on its similarity to the retained modality. These positive and negative pairs are processed by $\mathbf{\Phi}_c$ to generate multi-modal embeddings $\hat{\mathbf{f}}_c$. The class token from $\hat{\mathbf{f}}_c$ is then passed through a linear layer for matching prediction. The multi-modal matching process is optimized using the following binary cross-entropy loss: 
\begin{equation}
\mathcal{L}_{itm} = -\frac{1}{N} \left[
\sum_{i=1}^{N_1} \log(p_i^+) +
\sum_{j=1}^{N_2} \log(1 - p_j^-)
\right],
\end{equation}
where $N=N_1+N_2$, $N_1$ and $N_2$ denote the number of positive and negative pairs respectively. The terms $p_i^+$ and $p_j^-$ represent the predicted matching probabilities for the $i$-th positive pair and $j$-th negative pair. Overall, the final SSL objective is formulated as: $\mathcal{L} = \frac{1}{2}(\mathcal{L}_{itc} + \mathcal{L}_{itm})$.

\begin{table*}[!t]
\centering
\setlength{\tabcolsep}{1.2mm}
{
\begin{tabular}{c|ccc|ccc|ccc|ccc}
\Xhline{1pt}
\multirow{2}{*}{\textbf{Approaches}} & \multicolumn{3}{c|}{Attributes} & \multicolumn{3}{c|}{\textbf{ADNI}} & \multicolumn{3}{c|}{\textbf{AIBL}} & \multicolumn{3}{c}{\textbf{NACC}} \\
\cline{2-13}
 & \textit{I} & \textit{T} & \textit{S} & Acc & AUC & $\mathrm{F_{1}}$ & Acc & AUC & $\mathrm{F_{1}}$ & Acc & AUC & $\mathrm{F_{1}}$ \\
\hline
3D-ResNet-50 \cite{he2016deep} & \checkmark  &  &  & 0.578 & 0.743 & 0.536 & 0.720 & 0.776 & 0.440 & 0.668 & 0.726 & 0.466 \\
Transformer \cite{vaswani2017attention} &  & \checkmark &  & 0.799 & 0.936 & 0.797 & 0.801 & 0.820 & 0.556 & 0.715 & 0.825 & 0.595 \\
Concat-fusion \cite{spasov2019parameter} & \checkmark  & \checkmark &  &0.826 & \underline{0.952} & \underline{0.822} & 0.736 & 0.825 & 0.497 & 0.712 & 0.847 & 0.560 \\
DAFT \cite{wolf2022daft} & \checkmark  & \checkmark &  & 0.811 & 0.937 & 0.809 & 0.785 & 0.842 & 0.550 & 0.712 & 0.849 & 0.582 \\
Cross Attention \cite{zhang2023multi} & \checkmark  & \checkmark & & 0.829 & \underline{0.952} & 0.816 & 0.781 & 0.841 & 0.583 & 0.703 & 0.849 & 0.585 \\
Hyperfusion \cite{duenias2025hyperfusion} & \checkmark  & \checkmark & & 0.724 & 0.845 & 0.727 & 0.785 & 0.810 & 0.547 & 0.681 & 0.792 & 0.588\\
MMCL \cite{Hager_2023_CVPR} & \checkmark  & \checkmark & \checkmark & 0.599 & 0.769 & 0.591 & 0.703 & 0.714 & 0.420 & 0.653 & 0.751 & 0.453 \\
TIP \cite{du2024tip} & \checkmark  & \checkmark & \checkmark & 0.826 & 0.946 & 0.821 & \underline{0.793} & \underline{0.881} & \underline{0.628} & \underline{0.718} & \underline{0.850} & \underline{0.635} \\
ViTa \cite{zhang2025towards} & \checkmark & \checkmark & \checkmark & \underline{0.831} & 0.939 & 0.817 & 0.780 & 0.853 & 0.604 & \underline{0.718} & 0.844 & 0.634 \\
\textbf{Ours} & \checkmark  & \checkmark & \checkmark & \textbf{0.846} & \textbf{0.960} & \textbf{0.834} & \textbf{0.817} & \textbf{0.895} & \textbf{0.682} & \textbf{0.728} & \textbf{0.851} & \textbf{0.651} \\
\Xhline{1pt}
\multicolumn{13}{l}{\makecell[l]{$\ast$ \textit{I}, \textit{T} and \textit{S} refer to image modality, tabular modality and self-supervised learning.}} \\
\multicolumn{10}{l}{\makecell[l]{$\ast$ MMCL only adopt image modality during fine-tuning and inference stages.}} \\

\end{tabular}}
\caption{Quantitative comparison across multiple datasets. The best and second-best results are marked in \textbf{bold} and \underline{underline}.}
\label{tab:comparison}
\end{table*}

\section{Experiments}

\subsection{Experiment Settings}

\subsubsection{Datasets}
We collect data from three public brain imaging cohorts, including Alzheimer’s Disease Neuroimaging Initiative (ADNI) \cite{adni}, Australian Imaging, Biomarkers and Lifestyle (AIBL) \cite{aibl} and National Alzheimer’s Coordinating Center (NACC) \cite{nacc} cohorts. For imaging data, we collect T1-weighted MRIs and follow a standardized pipeline to pre-process these volumetric images. Specifically, we sequentially perform bias field correction \cite{tustison2010n4itk} and affine registration \cite{avants2014insight} to the MNI space \cite{fonov2009unbiased,fonov2011unbiased} with Clinica platform \cite{routier2021clinica}. Additionally, we perform skull-stripping via an open-source tool \cite{hoopes2022synthstrip}. 

For tabular data, following \cite{xue2024ai}, we manually collect 42 variables relevant to Alzheimer’s disease diagnosis cross 3 cohorts, including demographic information, genotype information 
and so on. 
We follow the time aligning procedure as \cite{zhang2024modality} for constructing image-tabular pairs. After acquiring the dataset, we convert all categorical variables to numerical values and normalize continuous variables using the mean and standard deviation of non-empty cells in corresponding column. Missing tabular values are assigned a value of 0. Finally, we obtain 1,689, 1,161 and 1,611 MRI-tabular pairs from ADNI, AIBL and NACC cohorts. The collected subjects encompass three types of cognitive degeneration, i.e., cognitive normal (CN), mild cognitive impairment (MCI) and dementia (DE). The pre-pocessed data are randomly split into training, validation and test set with the ratio of 7:1:2. To avoid data leakage, we perform a subject-wise split with no overlap across train, validation and test sets, and retain only one timepoint sample per subject.



\subsubsection{Implementation details}

We pretrained our model for 150 epochs on two NVIDIA
RTX A100 GPUs, with batch size of 12 on each gpu and learning rate of 0.0001. We adopt the Adam optimizer \cite{kingma2014adam} together with the weight decay of $1.5 \times 10^{-6}$. The pretraining was conducted on three collected dataset in cross-tabular manner. The temperature hyperparameter $\tau$ is set as 0.1. During fine-tuning stage, we initialized the model with the pretrained weight and train it on specific data cohort for 20 epochs with batch size of 12 on each gpu and learning rate of $1 \times 10^{-4}$. We retrain models which achieve the best performance over the validation set for evaluating the performance over test set.

\subsection{Comparison with State-of-the-art Methods}

In this section, we quantitatively compare our CITab method with 9 other state-of-the-art (SOTA) methods, i.e., ResNet-50 \cite{he2016deep}, Transformer \cite{vaswani2017attention}, Concat-fusion\cite{spasov2019parameter}, DAFT\cite{wolf2022daft}, Cross-attention fusion \cite{zhang2023multi}, Hyperfusion \cite{duenias2025hyperfusion}, MMCL \cite{Hager_2023_CVPR}, TIP \cite{du2024tip} and ViTa \cite{zhang2025towards}. These compared methods can be divided into 3 categories, i.e., uni-modal method (e.g., 3D-ResNet50), multi-modal fusion method (e.g., DAFT) and multi-modal SSL method (e.g., MMCL, TIP and ViTa).
We evaluate the performance in terms of 3 metrics, i.e., accuracy (Acc), area under the receiver operating characteristic curve (AUC) and $\mathrm{F_{1}}$-score. As shown in Table \ref{tab:comparison}, our CITab method outperforms other methods over all 3 datasets. For instance, our method outperforms the second-best method by 1.5\%, 2.4\% and 1.0\% in terms of Acc over ADNI, AIBL and NACC dataset.
Similar results can be observed on other two metrics. 

As shown in Table \ref{tab:comparison}, we can observe that methods relying solely on imaging modalities show significant performance degradation compared to multi-modal approaches. This is because diagnosing CN, MCI, and DE based only on MRI is inherently challenging, since MCI is a transitional stage, its imaging observation differences are too subtle to be distinguished from other two categories.  It is worth noting that MMCL only use imaging modality during its inference stage, leading to its inferior performance compared to other multi-modal methods. Generally, multi-modal SSL methods outperform multi-modal and uni-modal approaches. However, these models \cite{du2024tip, Hager_2023_CVPR} exhibit relatively limited performance gains, since they do not support cross-tabular learning and thus fail to capture the medical knowledge underlying heterogeneous tables. In contrast, our method achieves more performance gains than other multi-modal SSL methods, underscoring the necessity of scalable cross-tabular learning.

\begin{table*}[htbp]
\centering
\begin{tabular}{c|ccc|ccc|ccc}
    \Xhline{1pt}
    \multirow{2}{*}{\textbf{Settings} \textbf{configuration}} & \multicolumn{3}{c|}{\textbf{ADNI}} & \multicolumn{3}{c|}{\textbf{AIBL}} & \multicolumn{3}{c}{\textbf{NACC}}\\
    \cline{2-10}
    & Acc & AUC & $\mathrm{F_{1}}$ & Acc & AUC & $\mathrm{F_{1}}$ & Acc & AUC & $\mathrm{F_{1}}$ \\
    \hline
    \textit{w/o} multi-modal SSL & 0.828 & 0.939 & 0.817 & 0.756 & 0.790 & 0.498 & 0.699 & 0.800 & 0.579 \\
    
    \textit{w/o} header embedding & 0.654 & 0.862 & 0.608 & 0.752 & 0.793 & 0.496 & 0.690 & 0.791 & 0.547\\
    
    \textit{w/o} Mask Attention in $\mathbf{\Phi}_t$ and $\mathbf{\Phi}_c$ & 0.805  & 0.937 & 0.786 & 0.801 & 0.850 & 0.589 & 0.715 & 0.809 & 0.586 \\
    
    \textit{w/o} P-MoLin & 0.823 & 0.945 & 0.811 & 0.793 & 0.868 & 0.653 & \textbf{0.733} & 0.835 & 0.641 \\
    
    \textit{w/o} $\mathcal{L}_{itc}$ & 0.735 & 0.884  & 0.731 & 0.744 & 0.807 & 0.571 & 0.700 & 0.806 & 0.573 \\
    \textit{w/o} $\mathcal{L}_{itm}$ & 0.799 & 0.936 & 0.777 & 0.797 & 0.858 & 0.543 & 0.706 & 0.803 & 0.593 \\
    \hline
    Replace with Cross-attention  & 0.837 & 0.944 & 0.823 & 0.744 & 0.789 & 0.481 & 0.690 & 0.821 & 0.639 \\
    Replace with BioBERT embedding  & 0.834 & 0.939 & 0.815 & 0.736 & 0.814 & 0.474 & 0.706 & 0.806 & 0.618 \\
    \hline
    \textbf{Ours} & \textbf{0.846} & \textbf{0.960} & \textbf{0.834} & \textbf{0.817} & \textbf{0.895} & \textbf{0.682} & 0.728 & \textbf{0.851} & \textbf{0.651}  \\
    \Xhline{1pt}
\end{tabular}
\caption{Quantitative results of ablation study and comparison with alternative modules on three datasets.}
\label{tab:ablation}
\end{table*}

\subsection{Ablation Study}
\subsubsection{Contribution of key components}
We evaluate the contribution of key components in CITab through a set of ablation studies, where each component is removed individually to evaluate its effect on overall performance. The setting configurations include: 
(1) \textit{w/o} multi-modal SSL: we directly train the multi-modal method on each dataset without conducting SSL. (2) \textit{w/o} header embedding: we remove the header embedding. (3) \textit{w/o} mask attention: we replace mask attention with vanilla self-attention when modeling tables with various length. (4) \textit{w/o} P-MoLin module: we replace the P-MoLin module with vanilla linear layer.
(5) \textit{w/o} $\mathcal{L}_{itc}$: we remove the image-tabular contrastive loss during pretraining. (6) \textit{w/o} $\mathcal{L}_{itm}$: we remove the image-tabular matching loss during pretraining.  As shown in Table \ref{tab:ablation}, the performance decreases when independently removing any of the above components, showcasing their effectiveness. Notably, when removing header name embedding, the performance dramatically drops. This indicates that semantic comprehension is essential for modeling heterogeneous tabular data under cross-tabular scenarios.

\subsubsection{Comparison with alternatives}
In this section, we compare our method with alternative modules. As shown in Table \ref{tab:ablation}, in terms of multi-modal fusion process, we compare our method (unified self-attention) with cross-attention mechanism. The results indicate that our method outperforms the cross-attention mechanism, which implies that a unified self-attention architecture can model heterogeneous tabular data with differing numbers of columns more effectively. For header embedding, we employ BERT to embed the header names in this paper. Here we apply another language model BioBERT \cite{lee2020biobert}, which is specifically pretrained with medical corpus. As shown in Table \ref{tab:ablation}, using BERT for header embedding outperforms the BioBERT counterpart. To investigate this phenomenon, we computed the average token-wise similarity of the header embeddings generated by BERT and BioBERT. We found that the embeddings generated by BioBERT show a higher inter-token similarity, 0.935 across 3 dataset compared with 0.864 of BERT. BioBERT’s higher similarity stems from its pretraining on biomedical corpora, which enhances semantic closeness for domain-specific terms but results in more homogeneous representations. In contrast, BERT’s embeddings trained on general-domain text are more distinguishable, making them more effective for our model.

Additionally, we evaluate the performance of P-MoLin with varying numbers of expert pathways (prototypes). The quantitative results is tabulated in  Table \ref{tab:expert}. As shown, the model performance improves consistently with the increase of expert numbers, indicating that a larger set of experts empower the model to learn more robust and expressive representations. However, when the number of experts increases to 6, a performance drop is observed, which may be attributed to overfitting caused by the  model complexity. Therefore, we set $E$ as 5 in our default configuration.

\begin{table}[htbp]
\centering
\label{tab:low}
    \begin{tabular}{c|c|cccc}
    \Xhline{1pt}
    \multicolumn{2}{c|}{{\textbf{Settings}}}
      & $E$=3 & $E$=4 & $E$=5  & $E$=6\\
    \hline
    \multirow{3}{*}{\textbf{ADNI}} 
    & Acc & 0.817 & 0.828 & 0.846 & 0.837 \\
    & AUC & 0.935 & 0.950 & 0.960 & 0.947 \\
    & $\mathrm{F_{1}}$ & 0.796 & 0.815 & 0.834 & 0.825 \\
    \hline
    \multirow{3}{*}{\textbf{AIBL}} 
    & Acc & 0.768 & 0.801 & 0.817 & 0.809 \\
    & AUC & 0.861 & 0.883 & 0.895 & 0.885 \\
    & $\mathrm{F_{1}}$ & 0.628 & 0.587 & 0.682 & 0.667 \\
    \hline
    \multirow{3}{*}{\textbf{NACC}} 
    & Acc & 0.718 & 0.743 & 0.728 & 0.715 \\
    & AUC & 0.850 & 0.832 & 0.851 & 0.846 \\
    & $\mathrm{F_{1}}$ & 0.672 & 0.698 & 0.651 & 0.630 \\
    \hline
    \multirow{3}{*}{\textbf{Average}} 
    & Acc & 0.768 & 0.791 & \textbf{0.797} & 0.787 \\
    & AUC & 0.882 & 0.888 & \textbf{0.902} & 0.893 \\
    & $\mathrm{F_{1}}$ & 0.699 & 0.700 & \textbf{0.722} & 0.707 \\
    \Xhline{1pt}
\end{tabular}
\caption{Quantitative results of variants with varying numbers of experts over three datasets. The best average results are marked in \textbf{bold}. }
\label{tab:expert}
\end{table}

\begin{figure}[!ht]
\centering

\includegraphics[width=\linewidth]{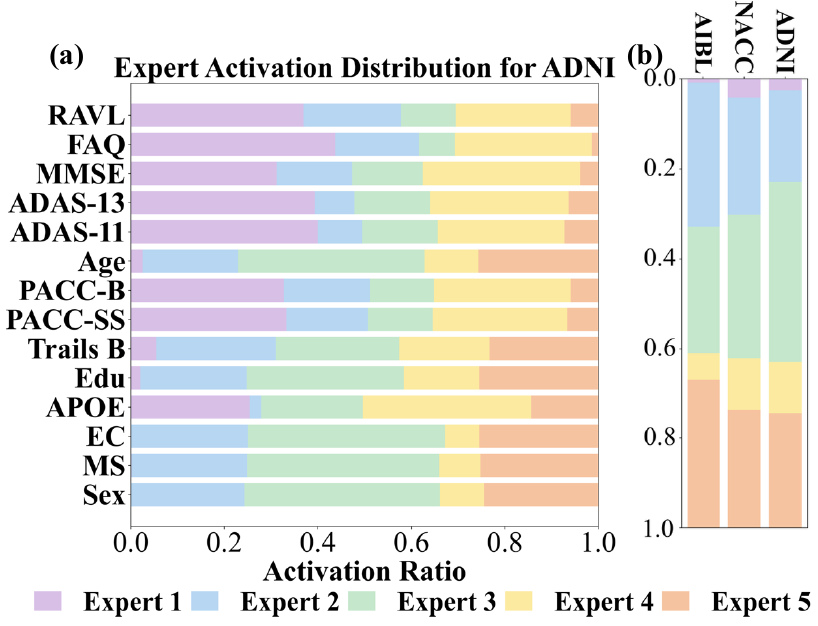}
\caption{Expert activation visualizations. (a) Ratios of experts with the highest activation values for different tabular features in ADNI dataset. (b) Ratios of experts with the highest activation values for the \textit{age} feature across three datasets.
}
\label{expert_ratio}
\end{figure}

\begin{figure}[!ht]
\centering

\includegraphics[width=.93\linewidth]{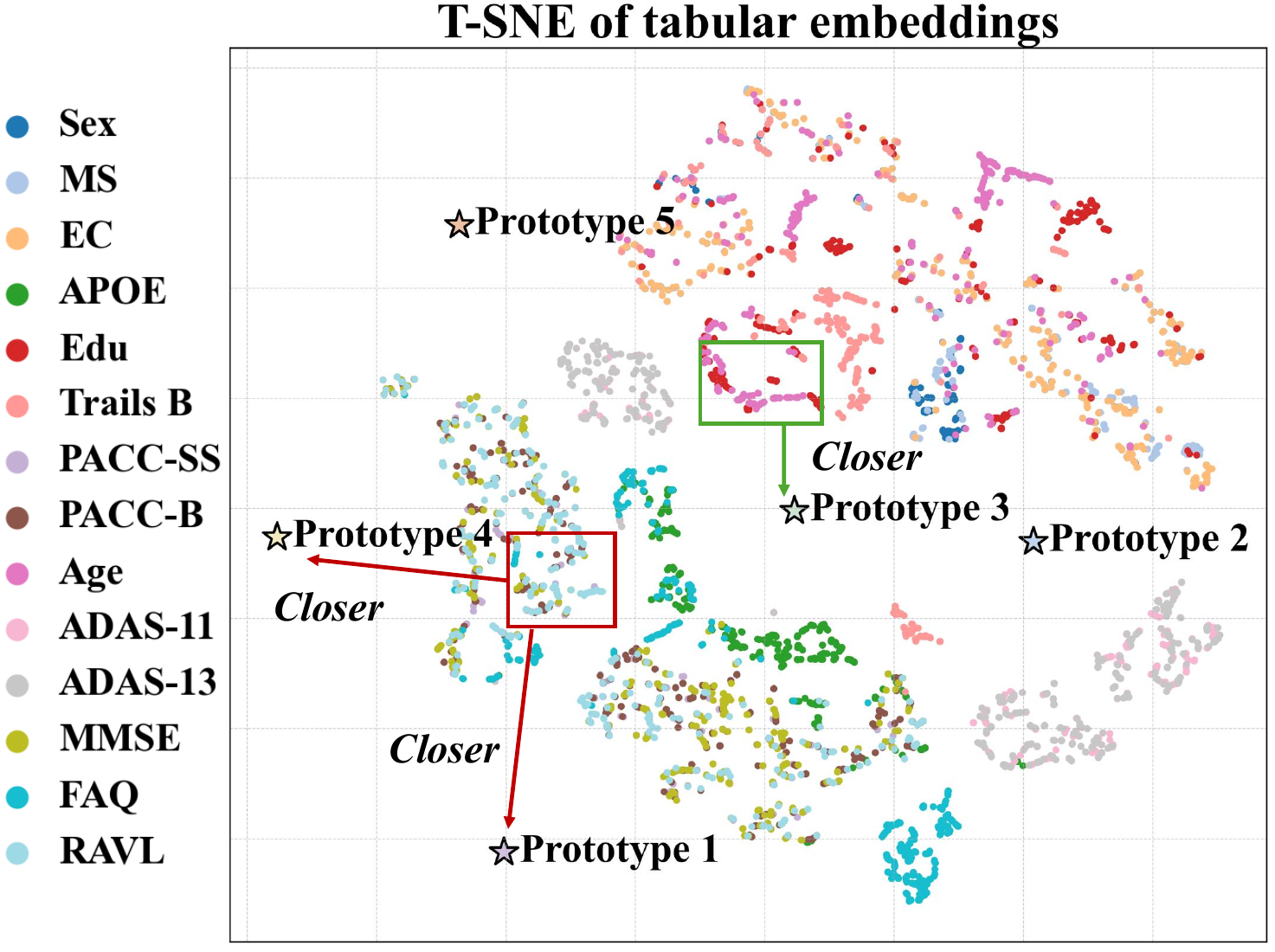}
\caption{T-SNE plots of the tabular feature embeddings and prototype embeddings of our method on ADNI dataset.
}
\label{prototype}
\end{figure}

\begin{figure}[!ht]
\centering
\includegraphics[width=.9\linewidth]{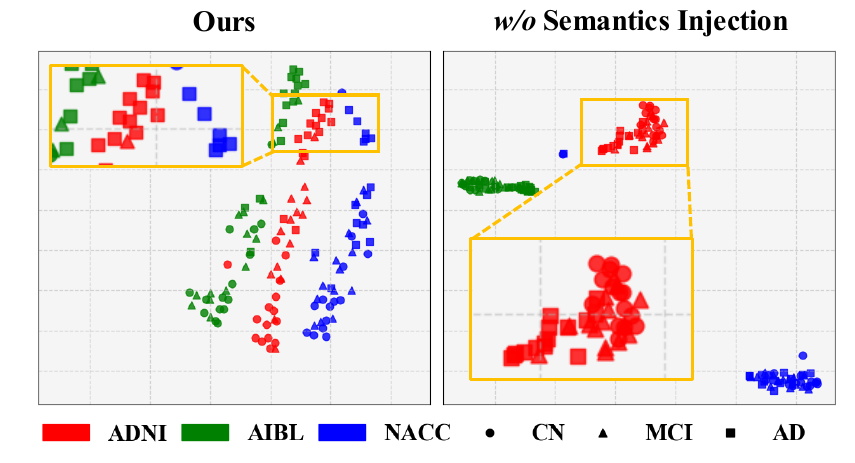}
\caption{
T-SNE plots of the feature embeddings from our proposed method with and without semantics injection. Different colors and shapes represent different data cohorts and disease types respectively.
}
\label{tsne}
\end{figure}

\subsection{Interpretable visualizations} 
\subsubsection{Expert activation visualization} 
We present a histogram depicting the ratios of expert indices corresponding to the predominantly activated FFNs in P-MoLin over ADNI dataset in Fig.\ref{expert_ratio}-(a). As depicted, the distribution differs across various features in the ADNI dataset, with different features tending to activate distinct experts. For example, the top rows related to cognitive abilities show a higher activation of Expert 1. Additionally, in Fig.\ref{expert_ratio}-(b), we plot the activation distribution of a selected shared feature, i.e., age. The activation distributions across the three datasets are quite similar. The above analysis demonstrates that the proposed P-MoLin module enhances the model's capability for feature specialization and knowledge generalization. 

\subsubsection{Tabular embedding visualization}
We utilize the t-SNE (t-distributed Stochastic Neighbor Embedding) method to visualize the tabular embeddings of our approach, depicting each token corresponding to a tabular feature and the learnable prototypes within our P-MoLin layer using the ADNI dataset. As shown in Fig.\ref{prototype}, the age feature cluster (highlighted by the green rectangle), which primarily activates Expert 3 and deactivates Expert 1 according to Fig.\ref{expert_ratio}, is closer to Prototype 3 in the embedding space and farthest from Prototype 1. Similarly, the RAVL feature cluster (marked by the red rectangle) shows similar evidence. Note that some prototypes lie outside the tabular feature clusters. This is mainly because we use these prototypes as anchors to measure relative similarities, without explicitly enforcing them to align with the tabular data manifold.

\subsubsection{Feature embedding visualization} We adopt the t-SNE method to visualize the multi-modal embeddings of our approach and the variant without semantic-awareness, i.e., removing the header embedding. 
As shown in Fig.\ref{tsne}, the samples with same disease from different datasets cluster more closely in the embedding space, showing the successful knowledge generalization. Meanwhile, the separation of different diseases within each dataset is more distinct, which shows the effectiveness of semantic-awareness.



\section{Conclusion}
In this paper, we propose CITab, a novel image-tabular SSL framework designed to learn transferable medical knowledge across distinct tables. To enhance semantic-awareness during tabular modeling, we introduce a header embedding mechanism to mitigate the challenges posed by heterogeneous tabular structures. Additionally, we propose a P-MoLin module for feature specialization. Experimental results demonstrate that our method outperforms other approaches, highlighting the effectiveness of cross-table learning in generating powerful image-tabular representations. 

\section{Acknowledgments}
This work was supported by Ministry of Education Tier 1 grant, Singapore (24-1250-P0001), and Ministry of Education Tier 2 grant, Singapore (T2EP20224-0028).


\bibliography{aaai2026}

\end{document}